# 3D LiDAR Aided GNSS NLOS Mitigation for Reliable GNSS-RTK Positioning in Urban Canyons

Xikun Liu, Weisong Wen*, Feng Huang, Han Gao, Yongliang Wang, Li-Ta Hsu

*Abstract*— GNSS and LiDAR odometry are complementary as they provide absolute and relative positioning, respectively. Their integration in a loosely-coupled manner is straightforward but is challenged in urban canyons due to the GNSS signal reflections. Recent proposed 3D LiDAR-aided (3DLA) GNSS methods employ the point cloud map to identify the non-line-of-sight (NLOS) reception of GNSS signals. This facilitates the GNSS receiver to obtain improved urban positioning but not achieve a sub-meter level. GNSS real-time kinematics (RTK) uses carrier phase measurements to obtain decimeter-level positioning. In urban areas, the GNSS RTK is not only challenged by multipath and NLOS-affected measurement but also suffers from signal blockage by the building. The latter will impose a challenge in solving the ambiguity within the carrier phase measurements. In the other words, the model observability of the ambiguity resolution (AR) is greatly decreased. This paper proposes to generate virtual satellite (VS) measurements using the selected LiDAR landmarks from the accumulated 3D point cloud maps (PCM). These LiDAR-PCM-made VS measurements are tightly-coupled with GNSS pseudorange and carrier phase measurements. Thus, the VS measurements can provide complementary constraints, meaning providing low-elevation-angle measurements in the across-street directions. As a result, the model observability of the AR is improved. The implementation is done using factor graph optimization to solve an accurate float solution of the ambiguity before it is fed into LAMBDA. The effectiveness of the proposed method has been validated by the evaluation conducted on our recently open-sourced challenging dataset, UrbanNav. The result shows the fix rate of the proposed 3DLA GNSS RTK is about 30% while the conventional GNSS-RTK only achieves about 14%. In addition, the proposed method achieves sub-meter positioning accuracy in most of the data collected in challenging urban areas.

## I. INTRODUCTION

High-precision positioning without prior information is one of the most crucial functionalities of the unmanned navigation system, including autonomous driving vehicles (ADV) [1], unmanned aerial vehicles (UAV) [2], etc. With more onboard sensors being available on autonomous systems, the integration of multiple sensors has been extensively explored over the past several decades. The sensor integration approaches have shown better performance and increased potential thanks to the complementation of the diverse characteristics of various sensors. The camera has always attracted research interest due to its informative perception and cost-effectiveness. The visual odometry (VO) and visual/inertial odometry (VIO) methods [3-7] have shown reliable results in most indoor scenes. However, researches in [8, 9] show that complex urban scenes still posed huge challenges to visual-based approaches due to the dynamic objects, varying illumination conditions, and continuous rapid movement. Compared with the camera, LiDAR sensors directly provide the accurate 3D measurement of point clouds, which is insensitive to changing illumination conditions. Recent state-of-the-art research on LiDAR odometry (LO) and LiDAR/inertial Odometry (LIO) [10-12] have already shown impressive performances in accurate real-time positioning and large-scale mapping, even in the highly dynamic environment [13]. Nevertheless, both camera-based and LiDAR-based odometry can only provide relative positioning. In other words, accumulated drift is inevitable under long-term operation.

The GNSS positioning [14] has been playing an indispensable role in many fields for decades, especially in the autonomous navigation system. As opposed to the camera- or LiDAR-based methods, the GNSS positioning provides drift-free results benefiting from the globally referenced measurements. The conventional GNSS single point positioning (SPP) methods reach meters-level accuracy based on pseudorange measurements due to the inherent errors from the atmosphere and clock systems [14]. The GNSS real-time kinematic (RTK) positioning [15] further explores the carrier phase measurements to provide centimeter-level positioning accuracy with the help of the corrections from reference stations in open-sky areas. To achieve this, the GNSS-RTK firstly estimates the float solution based on double-differenced (DD) [16] GNSS raw measurements (pseudorange and carrier phase) and then performs the integer least-squares algorithm (e.g LAMBDA [17]) based on the estimated float solution to resolve the integer ambiguity. However, the performance of GNSS-RTK can be significantly degraded in the urban canyon due to signal blockage and reflection by the surrounding environment, which is shown in Figure 1. There are two main factors leading to the impact. First (*challenge 1*), a considerable part of received GNSS raw measurements in highly urbanized areas is so-called non-line-of-sight (NLOS) [18], which denotes the satellite signal received by reflection as the satellite is blocked by buildings [18] or dynamic objects (e.g., double-decker

Xikun Liu, Weisong Wen, Feng Huang and Li-Ta Hsu are all with Hong Kong Polytechnic University, Hong Kong (correspondence e-mail: welson.wen@polyu.edu.hk).

Han Gao and Yongliang Wang are with Huawei Technologies (e-mail: gaohan14@huawei.com, wangyongliang775@huawei.com).

buses) [19]. Such polluted measurements can significantly degrade the accuracy of the float solution. Second (***challenge 2***), the buildings and dynamic objects in the urban canyon block a huge number of signals from satellites, resulting in a limited satellites' number. In other words, only satellites with high elevation angles are received. Such a case is called the *poor geometry distribution* [20], which limits the success rate of integer ambiguity resolution (AR) and the positioning accuracy as well.

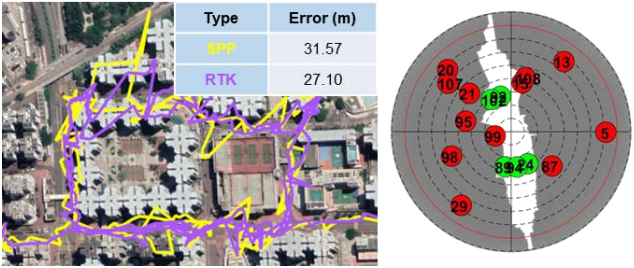

Fig. 1. The illustration of the challenges faced by the GNSS-RTK in an urban canyon. The left figure shows the comparison of trajectory and error in meters between SPP and GNSS-RTK positioning methods. The right figure shows the skyplot of a randomly selected location where the red and green circles denote the NLOS and LOS signals, respectively. The numbers indicate pseudorandom noise (PRN), which is used to distinguish different satellites.

Since the GNSS NLOS receptions are caused by the reflections from surrounding buildings, our previous work continuously proposed the LiDAR-aided GNSS SPP whereby the 3D LiDAR is utilized to reconstruct the environment to exclude [19, 21] or even correct [18, 22] the GNSS NLOS receptions. However, only an accuracy of several meters is obtained using the GNSS SPP. In our recent conference paper [23], we extended the 3D LiDAR-aided NLOS exclusion to the GNSS-RTK positioning. In particular, the GNSS NLOS is detected using the real-time 3D point cloud map (PCM) and excluded from the GNSS-RTK positioning. Unfortunately, the fixed solution is still difficult to obtain due to the poor satellite geometry in urban canyons. Moreover, the GNSS NLOS exclusion can further enhance this challenge. To fill this gap, this paper proposes a 3D LiDAR-aided (3DLA) GNSS-RTK positioning method which tackles the two listed dominant challenges. Specifically, the GNSS NLOS reception is detected and excluded through the incrementally generated and drift-free 3D sliding window PCM. Then, the sparsely selected LiDAR landmarks are used to create the so-called virtual satellite (VS) that improves the geometry distribution of the received standalone GNSS signals by tightly integrating the raw measurements from GNSS, VS, and inertial measurement unit (IMU). After solving the above optimization problem, AR is performed based on the float solution and covariance information improved by the virtual satellites to achieve the decimeter-level accuracy. Meanwhile, the improved float solutions and the fixed solutions are further employed to correct the drift of the 3D sliding window PCM for reliable NLOS detection. The main contributions of the proposed work are listed as follows:

1) *Tackle the first challenge of GNSS-RTK by GNSS NLOS mitigation*: This paper proposes a drift-free 3D sliding window PCM-based GNSS NLOS mitigation method to improve the quality of the raw GNSS measurements. The drift of the 3D sliding window PCM is corrected with the help of improved GNSS-RTK positioning.

2) *Tackle the second challenge of GNSS-RTK by geometry distribution improvement using virtual satellite*: This paper proposes a VS constraint model based on LiDAR landmarks to improve the geometry distribution of the received satellites in urban canyons. The improved float solution estimation can be obtained by tightly integrating the raw GNSS measurements, IMU, and carefully selected VS measurements, and then used for the integer ambiguity resolution. Moreover, we mathematically derive the geometry improvement arising from the virtual satellites.

3) *Extensive evaluation in urban canyons of Hong Kong*: Experiments have been conducted on two challenging urbanized sequences collected in Hong Kong and comprehensive comparisons are carefully conducted to show the effectiveness of the proposed method.

To the best of our knowledge, this is the first attempt to solve both the two key challenges in urban GNSS-RTK simultaneously through complementary LiDAR measurements. The rest of the paper is organized as follows: Section II introduces recent research on the listed two key problems of GNSS-RTK. The proposed 3D LiDAR-aided GNSS-RTK method with the detailed measurements model and residual formulation are given in Section III. Section IV elaborates on the experiments and evaluation results of the proposed method. Conclusion and potential further development are given in Section V. The notations and definitions are given in the Appendix.

## II. RELATED WORKS

This section reviews the related work on the two mentioned key challenges of GNSS-RTK positioning in urban canyons.

### A. GNSS NLOS Mitigation for Urban GNSS-RTK Positioning

***Consistency check aided GNSS outlier mitigation***: An intuitive method for detecting polluted GNSS outlier measurements is the use of measurements redundancy by consistency. The work in [24] performed outlier exclusion by consistency check. However, in deep urban areas where a very limited number of satellites are received, the performance of the consistency check method can be limited by the lack of healthy measurements. With more onboard sensors being available in recent years, [25, 26] proposed GNSS, IMU, and camera-integrated systems through the extended Kalman filter (EKF) and performed the innovation-based outlier-rejection for GNSS measurements using IMU and camera predictions. As a result, the constraints from healthy GNSS measurements together with IMU and camera measurements jointly contributed to obtaining a better float solution and reduced the searching space of AR. [27] presented a tightly-coupled filtering-based GNSS precise point positioning (PPP)/INS/LiDAR integrated system, similarly utilizing the consistency between IMU and GNSS measurements to reject gross outliers. Nevertheless, such an outlier-resistant strategy relies on the accuracy of INS and the heuristically determined threshold. Also, [28] proposed LiDAR-aided GNSS receiver autonomous integrity monitoring (RAIM) that performed outlier exclusion through the residual consistency check on the fused position. However, it requires an accurate initial guess of the state for reliable outlier detection.

*Environmental perception aided GNSS NLOS mitigation*: Given that NLOS receptions could have the most negative impact on GNSS-RTK positioning methods in urban canyons [18], research on NLOS detection and exclusion is widely conducted. The work in [29] utilized 3D building models and prior positions to assess signal quality, the potential GNSS NLOS was removed from further GNSS-RTK positioning. The work in [30] further proposed to use 3D model-aided multi-hypothesis method to improve the GNSS-RTK positioning in urban canyons. NLOS exclusion and GNSS-RTK positioning are performed and evaluated on each hypothetical initial position and the final result is achieved by the weighted average of the estimated hypothesis. However, it is difficult to obtain an accurate initial guess and precise environmental model for successful NLOS detection during real-time positioning. Therefore, researchers proposed to exploit the active-perception-based NLOS mitigation methods that take the advantage of the onboard sensor (e.g., camera and LiDAR). The work in [31] utilized a sky-pointing fish-eye camera with super-pixel segmentation to classify open sky areas and blocked sky areas by buildings. Afterward, the visibility of the received satellites can be used for NLOS detection and exclusion. Furthermore, [32] also utilized the fish-eye camera to identify the NLOS signals, which were then adaptively weighted instead of directly excluded for further tightly-coupled GNSS/INS integration. The result showed better performance as the geometry distribution of the available satellites is not deteriorated by avoiding signal exclusion. Different from [32], [33] proposed using LiDAR for providing more environmental information to correct the NLOS measurements after visual-aided NLOS detection. Nevertheless, these camera-aided GNSS NLOS detection methods are sensitive to illumination conditions. On the other side, our previous research on LiDAR-aided NLOS detection has also shown effectiveness through the improvement in positioning accuracy as introduced in Section I. However, those methods [18, 19, 21, 22] detected the NLOS receptions based on either the single frame point cloud, which is limited by the narrow FOV of LiDAR, or the accumulated multiple frame point clouds by the state-of-the-art LIO methods, which are subject to drift over time. How to reliably and complementarily integrate the GNSS and LiDAR for GNSS NLOS mitigation remains to be exploited, which is one of the key contributions of this paper.

### B. Geometry Distribution Improvement for Urban GNSS-RTK Positioning

Apart from the unhealthy measurements, the poor satellite geometry distribution is another factor limiting the performance of the GNSS-RTK in urban canyons. In particular, poor satellite geometry can lead to significantly increased ambiguity dilution of precision (ADOP) [34]. As a result, the integer ambiguity of the carrier phase is hard to be reliably resolved. Improving the geometry constraint of the GNSS-RTK is a new research area. The work in [35] proposed a fuzzy LiDAR feature-matching method to constrain the rover's position based on the prior feature map. Then the original GNSS-only constraints were augmented as the LiDAR-aided GNSS constraints. The least-squares method was adopted to solve the augmented positioning problem and a higher fix rate was found. [36] extracted LiDAR features via the deep learning method and construct the absolute LiDAR constraints by matching the selected features to a pre-defined high-definition (HD) map. The mix constraint model including GNSS and LiDAR was solved by the weighting least-squares method. [36] also performed a comprehensive evaluation to indicate the effectiveness of the LiDAR-aided AR. However, [35, 36] both relied on prior maps for providing absolute LiDAR constraints, which is hardly guaranteed in actual applications. Moreover, without outlier exclusion, the quality of the initial guesses can be very poor, which results in unsuccessful feature associations. [37] integrated the LiDAR and GNSS constraints through the extended Kalman filter, which enhanced the original geometric distribution. In particular, the work assumed that the planar surfaces are repeatedly observable within multiple frames of 3D point clouds which is hard to accomplish in complex urban canyons. It also applies a parallel particle filter (PF) in the position domain to alleviate the impacts from cycle slip, NLOS, or multipath receptions. Nevertheless, in deeply urbanized areas, most of the received signals can be polluted by the aforementioned effects, which brings great limitations as most of the GNSS measurements are highly biased that cannot be used directly, and the filter-based method cannot fully make use of the time-correlation within the historical information simultaneously [38].

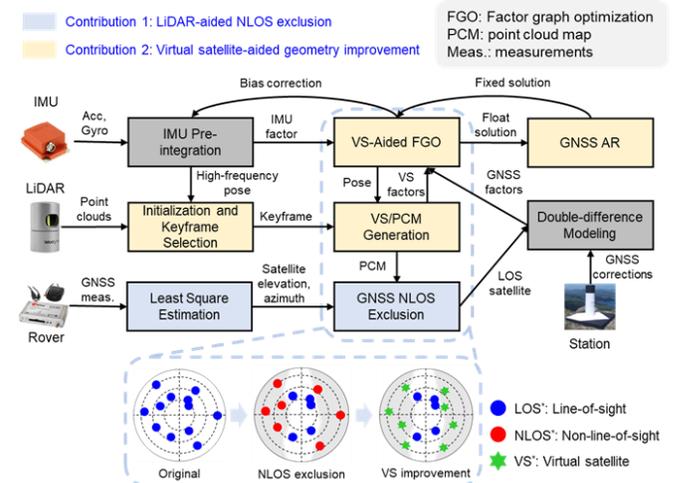

Fig. 2. The overview of the system pipeline.

## III. METHODOLOGY

### A. System Overview

The pipeline of the proposed system is depicted in Figure 2. The system consists of two major parts, namely LiDAR-aided NLOS exclusion, and integrated positioning with VS-aided geometry improvement.

The inputs of the system contain raw GNSS measurements (pseudorange measurement, carrier phase measurement, and Doppler measurement), LiDAR measurements, and IMU measurements. All the measurements are processed under a common keyframe mode [3] to keep the balance between computational load and information redundancy. Firstly, the pre-integration [39] technique is used to preprocess the raw IMU measurements. Meanwhile, the initial pose of each new keyframe in the sliding window is propagated through the IMU measurements. Then the poses of the keyframes and the correspondingly generated local 3D PCM are utilized to detect the NLOS and cycle slip [40] receptions. Meanwhile, the VS constraints are created from the extracted features and PCM [10]. After preprocessing, the states in the sliding window are jointly optimized through factor graph optimization (FGO) based on the constraints from the VS, GNSS, and IMU to obtain the improved float solution, which includes the positioning result and the covariance information. Afterward, the optimized position and the covariance matrices are further applied for AR to find fix solution. Finally, the global pose graph optimization is performed based on the fixed and float solutions to get the final pose result. The updated global poses, as well as the corresponding PCM, will further contribute to further GNSS NLOS detection. The marginalization strategy [3] is additionally employed to ensure real-time performance by marginalizing the measurements outside the sliding window. The detail of the methodology is introduced in the following Section III-B and Section III-C.

### B. 3D LiDAR-Aided GNSS NLOS Mitigation

For the new coming GNSS measurements from the satellites, NLOS detection and exclusion are first performed. In this paper, we employ a similar fast-searching method developed in our previous work [20] to classify the satellite visibilities based on the local 3D PCM. The 3D PCM is accumulated based on the point clouds of recent keyframes. Compared with our previous method in terms of the GNSS NLOS exclusion, the proposed method in this paper alleviates the drift of the 3D PCM by utilizing the corrections from the GNSS-RTK. Although the satellite visibility classification based on the 3D PCM is not new, this paper still briefly sketches the key steps for completeness.

Our method contains three major steps: ***First***, a sliding window-based local PCM, which is accumulated by last $n_{NL}$ keyframes, is simultaneously maintained. $n_{NL}$ is determined for PCM distance reaching approximately 250m, as GNSS NLOS is primarily caused by objects within a certain distance, according to [18]. ***Second***, as shown in Figure 3a, an orientation-based fixed-step search is applied on the local PCM to classify the visibility of the satellite. To be more specific, we calculate the line-of-sight (LOS) vector based on current positions of receiver and satellite. Then, we set the search point to the receiver's current position $\mathbf{p}_{r,t}^{EN}$ and move it in steps of $\Delta d$ along the LOS vector and check the number of the neighboring map points within the search range $\Delta r$. If the neighboring number is larger than the threshold, then the satellite will be classified as invisible, and the relevant measurement is excluded as NLOS receptions. However, when the drift is accumulated during vehicle driving, the attitude error subsequently brings great bias to the local map direction, which leads to inaccurate NLOS classification. In other words, the 3D PCM-aided NLOS detection relies heavily on accurate attitude estimation, which is easily affected by the drift. Figure 3b illustrates how the drifted PCM will affect the detection result. Therefore, we further propose, in contrast to our previous approach in [20], to apply the global pose graph optimization to update the local PCM simultaneously against the potential drift error, which enables better NLOS detection by making it free from attitude bias. The global optimization is based on the improved GNSS-RTK positioning results introduced in the following sections. Figure 3b also shows the illustration of the proposed drift-free PCM-based NLOS exclusion. It is observed that, before LiDAR-aided NLOS exclusion, near one-third of the satellites blocked by buildings and trees are originally used as LOS satellites, which causes large positioning errors. On the other hand, the drift-free PCM provides the correct environmental information, which leads

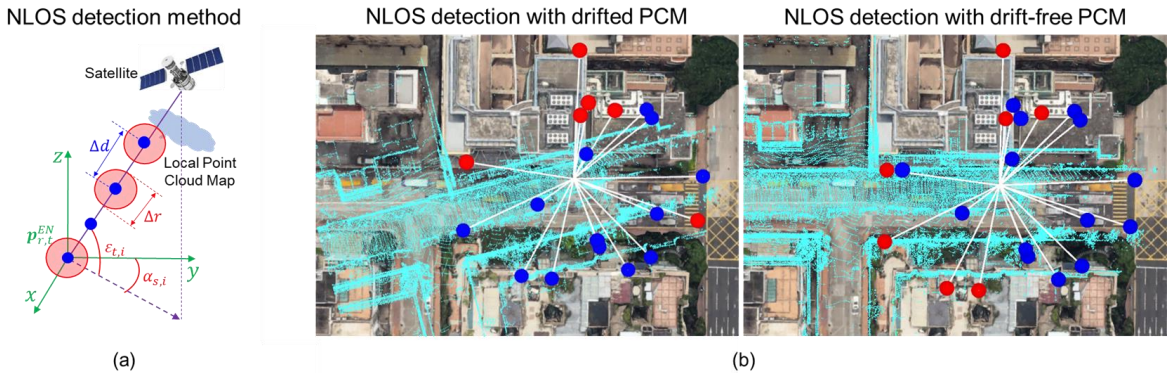

Fig. 3. Illustration of the proposed 3D LiDAR-aided NLOS detection based on drift-free PCM. Figure (a) shows the detail of the detection method [20]. Figure (b) shows the comparison of satellites' occlusion in the same epoch by drifted- and drifted-free-PCM. The red dots are detected NLOS satellites, while the blue dots represent the LOS satellites. The detection result and the PCM are projected on *Google Earth*.

to accurate NLOS detection.

### C. Virtual Satellite aided GNSS-RTK/IMU Factor Graph Optimization and Ambiguity Resolution

In this section, the method of tightly-coupled VS-aided GNSS-RTK/IMU FGO and AR is introduced, namely the modeling of the measurements, the construction of the integrated factor graph as well as the derivation of VS-aided AR. The system states are initialized in the ENU frame based on extrinsic parameters between different sensors. Notably, the body frame is aligned with the center of IMU. All the involved system states are listed as:

a) The position $\mathbf{p}_{b,k}^{EN}$ and orientation $\mathbf{q}_{b,k}^{EN}$ of the IMU in ENU coordinate. $k$ represents the $k^{th}$ keyframe in the sliding window.

b) The velocity $\mathbf{v}_{b,k}^{EN}$ of the IMU in ENU coordinate. $\mathbf{b}_a$ and $\mathbf{b}_w$ represent the bias of the gyroscope and accelerometer, respectively.

c) The DD integer ambiguities $N_{DD,r,t}^{S}$ of all received satellite $s \in S$ to receiver $r$ at time epoch $t$.

d) The clock drift $\dot{\delta}_{r,t}$ of receiver $r$ at time epoch $t$.

The system states within the sliding window can be further expressed as:

$$\chi = [x_0, \cdots, x_{K-1}, N_{DD,r,t_0}^{S}, \cdots, N_{DD,r,t_{n-1}}^{S}, \dot{\delta}_{t_0}^{r}, \cdots, \dot{\delta}_{t_{n-1}}^{r}] \quad (1)$$

$$\text{With } x_k = [\mathbf{p}_{b,k}^{EN}, \mathbf{q}_{b,k}^{EN}, \mathbf{v}_{b,k}^{EN}, \mathbf{b}_a, \mathbf{b}_w]$$

$$N_{DD,r,t}^{S} = [N_{DD,r,t}^{s_0}, \cdots, N_{DD,r,t}^{s_{m-1}}]$$

where $k \in [0, \cdots, K-1]$, $K$ represents the size of the sliding window, $t \in [t_0, \cdots, t_{n-1}]$ represents the received epoch of the GNSS signal, $n$ represents the epoch number of the received GNSS measurements within the time interval of the sliding window, and $m$ denotes the number of received satellites at time epoch $t$.

To obtain the optimal state estimation based on the given measurements, the maximum posterior probability should be reached. In this paper, the measurements are regarded as independent and with zero-mean Gaussian-distributed noise. The problem can be further simplified as solving the following objective function:

$$\chi^* = \underset{\chi}{\operatorname{argmin}} \sum_{S,r,k,t} (\|\mathbf{r}_p - \mathbf{H}_p\chi\| + \|\mathbf{r}_{L,k}\|_{\Sigma_L}^{2} + \|\mathbf{r}_{B,k}\|_{\Sigma_B}^{2} +$$

$$\|\mathbf{r}_{DD,\rho,r,t}^{S}\|_{\sigma_\rho}^{2} + \|\mathbf{r}_{DD,\psi,r,t}^{S}\|_{\sigma_\psi}^{2} + \|\mathbf{r}_{DD,N,r,t}^{S}\|_{\sigma_N}^{2} + \|\mathbf{r}_{d,r,t}^{S}\|_{\sigma_d}^{2}) \quad (2)$$

where $\{\mathbf{r}_p, \mathbf{H}_p\}$ denotes the marginalized term as prior constraints. $\mathbf{r}_{B,k}$ represents the IMU factor, which is weighted by the relative covariance matrix $\Sigma_B$. Similarly, $\mathbf{r}_{L,k}$ represents the VS factor and is weighted by the covariance matrix $\Sigma_L$. $\mathbf{r}_{DD,\rho,r,t}^{S}$, $\mathbf{r}_{DD,\psi,r,t}^{S}$ and $\mathbf{r}_{d,r,t}^{S}$ denote the DD GNSS pseudorange, carrier phase, and Doppler factors, respectively. The relative covariance matrices are with different weighting as $\sigma_\psi = \frac{\sigma_\rho}{100}$ and $\sigma_d = \sigma_\rho$, where the $\sigma_\rho$ is initially calculated by SNR and elevation angle from [41]. $\mathbf{r}_{DD,N,r,t}^{S}$ denotes the constant integer ambiguity factor weighted by $\sigma_N = \sigma_\psi$.

The factor graph structure of the proposed system is shown in Figure 4. It should be noted that the VS factors and IMU factors directly constraint the system states $x_k$. Differently, due to the difference in the data frequency, the GNSS factors (DD pseudorange, DD carrier phase, and Doppler factors) constrain the system states $x_k$ and $x_{k+1}$ through interpolated states $x_t$ at time epoch $t$, with $x_t \in (x_k, x_{k+1})$.

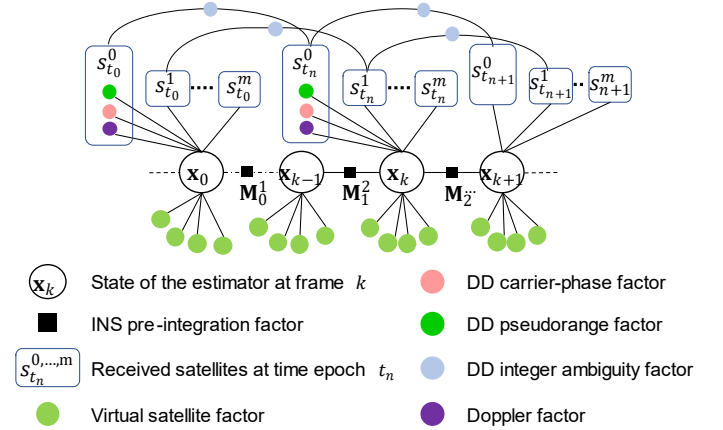

Fig. 4. The factor graph of the proposed tightly-coupled integration system.

#### 1) GNSS DD Pseudorange/Carrier phase Factor

The pseudorange measurement $\rho_{r,t}^{s}$ of the GNSS receiver $r$ at time $t$ is generally expressed by [42]:

$$\rho_{r,t}^{s} = r_{r,t}^{s} + c(\delta_{r,t} - \delta_{s,t}) + I_{r,t}^{s} + T_{r,t}^{s} + \varepsilon_{\rho,r,t}^{s} \quad (3)$$

where $r_{r,t}^{s}$ denotes the real geometric range between the satellite $s$ and receiver $r$ in time epoch $t$. $I_{r,t}^{s}$ represents the ionospheric delay distance, $T_{r,t}^{s}$ represents the tropospheric delay distance. $\varepsilon_{\rho,r,t}^{s}$ denotes the rest errors including multipath error, NLOS error, receiver noise error, and antenna phase-related noise error.

Similar to the pseudorange, the carrier phase measurements of the GNSS receiver $r$ at time $t$ can be expressed as [42]:

$$\lambda\psi_{r,t}^{s} = r_{r,t}^{s} + c(\delta_{r,t} - \delta_{s,t}) - I_{r,t}^{s} + T_{r,t}^{s} + \lambda B_{r,t}^{s} + d\psi_{r,t}^{s} + \varepsilon_{\psi,r,t}^{s} \quad (4)$$

where $B_{r,t}^{s} = \psi_{r,t,0} - \psi_{0,r,t}^{s} + N_{r,t}^{s}$ represents the carrier phase bias. The variable $\psi_{r,t,0}$ denotes the initial phase of the receiver's local oscillator. $\psi_{0,r,t}^{s}$ is the initial phase of the transmitted navigation signal from the satellite. The variable $N_{r,t}^{s}$ is the carrier phase integer ambiguity which should be an integer value. $\lambda$ denotes the carrier wavelength of the respective satellite system. $d\psi_{r,t}^{s}$ represents the carrier phase correction terms, which contain antenna phase offsets and variations, station displacement by earth tides, phase windup effect, and relativity correction on the satellite clock. $\varepsilon_{\psi,r,t}^{s}$ represents the errors relative to multipath effects, NLOS receptions, receiver noise, and antenna delay.

Based on the measurements model, it is observed that the clock-based and atmosphere-based systematic errors have great influences on positioning accuracy. Therefore, the DD technique is introduced in the GNSS-RTK positioning method. The DD method is to first conduct a single-difference between the measurements by different receivers (receiver $r$ and reference station $e$) from the same satellite and then perform a subtraction between the results of the single difference from two satellites. Particularly, master satellite $w$ is selected with the highest elevation angle among all received satellites, as satellites with higher elevation angles are inclined to suffer less from multipath and NLOS receptions. For certain time epochs and satellite systems, all the other satellites share the same master satellite. The formulation of the DD pseudorange and DD carrier phase can be concluded as [42]:

$$\rho_{DD,r,t}^s = (\rho_{r,t}^s - \rho_{e,t}^s) - (\rho_{r,t}^w - \rho_{e,t}^w) \quad (5)$$

$$\psi_{DD,r,t}^s = (\psi_{r,t}^s - \psi_{e,t}^s) - (\psi_{r,t}^w - \psi_{e,t}^w) \quad (6)$$

Considering that the receiver $r$ and reference station $e$ are under similar atmosphere condition, a single difference operation can eliminate the effect of atmosphere error as well as the satellite clock bias, but the receiver clock bias term remain. The second difference operation further eliminates the receiver clock bias. Therefore, the DD pseudorange and DD carrier phase measurements model can be further expressed as:

$$\rho_{DD,r,t}^s = (r_{r,t}^s - r_{e,t}^s) - (r_{r,t}^w - r_{e,t}^w) + \varepsilon_{DD,\rho,r,t}^s \quad (7)$$

$$\lambda \psi_{DD,r,t}^s = (r_{r,t}^s - r_{e,t}^s) - (r_{r,t}^w - r_{e,t}^w) + N_{DD,r,t}^s + \varepsilon_{DD,\psi,r,t}^s \quad (8)$$

where $\varepsilon_{DD,\rho,r,t}^s$, $\varepsilon_{DD,\psi,r,t}^s$ represents the noise of DD pseudorange measurements and carrier phase measurements. $N_{DD,r,t}^s$ is the DD integer ambiguity of satellite $s$, which is one of the system states to be estimated.

Given the DD measurement model above, the DD pseudorange residual and DD carrier phase residuals are formed as:

$$r_{DD,\rho,r,t}^s = \rho_{DD,r,t}^s - (r_{r,t}^s - r_{e,t}^s) - (r_{r,t}^w - r_{e,t}^w) \quad (9)$$

$$r_{DD,\psi,r,t}^s = \lambda_i \psi_{DD,r,t}^s - (r_{r,t}^s - r_{e,t}^s) - (r_{r,t}^w - r_{e,t}^w) - N_{DD,r,t}^s \quad (10)$$

where the range distances $r_{r,t}^s$, $r_{e,t}^s$, $r_{r,t}^w$ and $r_{e,t}^w$ are calculated based on the positions of the GNSS receiver:

$$r_{r,t}^s = \|\mathbf{p}_{r,t}^{EC} - \mathbf{p}_{s,t}^{EC}\|, r_{e,t}^s = \|\mathbf{p}_e^{EC} - \mathbf{p}_{s,t}^{EC}\|,$$

$$r_{r,t}^w = \|\mathbf{p}_{r,t}^{EC} - \mathbf{p}_{w,t}^{EC}\|, r_{e,t}^w = \|\mathbf{p}_e^{EC} - \mathbf{p}_{w,t}^{EC}\|$$

where $\|*\|$ denotes the norm of the vector. $\mathbf{p}_{s,t}^{EC}$ and $\mathbf{p}_{w,t}^{EC}$ are satellite position transformed in the ECEF frame, $\mathbf{p}_e^{EC}$ is the position of the reference station in the ECEF frame. The estimated $\mathbf{p}_{r,t}^{EC}$ is the position of the GNSS receiver $r$ at time epoch $t$ in the ECEF frame transformed from $\mathbf{p}_{r,t}^{EN}$. Notably, the transformation from ENU to ECEF according to the origin point $\mathbf{p}_o^{EC}$ is calculated as:

$$\mathbf{p}_{r,t}^{EC} = \mathbf{R}_{EN}^{EC} \mathbf{p}_{r,t}^{EN} + \mathbf{p}_o^{EC} \quad (11)$$

$$\mathbf{R}_{EN}^{EC} = \begin{bmatrix} -\sin\lambda_o & -\sin\phi_o \cos\lambda_o & \cos\phi_o \cos\lambda_o \\ \cos\lambda_o & -\sin\phi_o \sin\lambda_o & \cos\phi_o \sin\lambda_o \\ 0 & \cos\phi_o & \sin\phi_o \end{bmatrix} \quad (12)$$

where $\lambda_o$ and $\phi_o$ denote the geographic latitude and longitude of the priorly known origin point $\mathbf{p}_o^{EC}$.

Further, the receiver's position $\mathbf{p}_{r,t}^{EN}$ is obtained from the estimated states $\mathbf{p}_{b,t}^{EN}$ maintained in the body frame as:

$$\mathbf{p}_{r,t}^{EN} = \mathbf{p}_{b,t}^{EN} + \mathbf{R}_r^b \mathbf{p}_r^b \quad (13)$$

In addition, the maintained states are with LiDAR keyframe time $t_k$ rather than GNSS epoch time $t$. Therefore, linear interpolation is adopted between the system states $\mathbf{p}_{b,k}^{EN}$ and $\mathbf{p}_{b,k+1}^{EN}$ in the adjacent moment $t_k$ and $t_{k+1}$ with $t \in [t_k, t_{k+1}]$ to obtain the correspondent state $\mathbf{p}_{b,t}^{EN}$, which is calculated based on the ratio of the time interval:

$$\mathbf{p}_{b,t}^{EN} = \left\{ \frac{t - t_k}{t_{k+1} - t_k} \mathbf{p}_{b,k}^{EN} + \frac{t_{k+1} - t}{t_{k+1} - t_k} \mathbf{p}_{b,k+1}^{EN} \right\} \quad (14)$$

*2) GNSS Constant Integer Ambiguity Factor*

In the application of GNSS-RTK in urban, the positioning accuracy would suffer if cycle slips are not detected and properly handled. Cycle slip occurs when the receiver's phase lock on the signal is lost. In urban canyons, the most common reason for the cycle slip is an obstruction (e.g., buildings, trees) which blocks the signal and therefore results in signal tracking failure. In this case, the previously resolved integer ambiguities become instantly unknown and need to be resolved again. When there is no cycle slip, the integer ambiguity of one satellite in adjacent epochs should remain the same. The constant integer ambiguity residual can be formed as:

$$r_{DD,N,r,t}^s = N_{DD,r,t}^s - N_{DD,r,t-1}^s \quad (15)$$

where $N_{DD,r,t}^s$ and $N_{DD,r,t-1}^s$ represent the integer ambiguities of satellite $s$ in time epoch $t$ and epoch $t-1$, respectively.

To effectively eliminate the impact of cycle slip, we adopt the LiDAR-aided cycle slip detection method in [40], which employs the consistency check of the triple difference (TD). TD is between two double differences over two different epochs. To perform TD estimation, we predict the states by LiDAR and IMU measurements. According to equation (8), the DD integer ambiguity for satellite $s$ in time epoch $t$ is estimated as:

$$N_{DD,r,t}^s = \lambda \psi_{DD,r,t}^s - \left((r_{r,t}^s - r_{e,t}^s) - (r_{r,t}^w - r_{e,t}^w)\right) \quad (16)$$

then the TD integer ambiguity between time epoch $t$ and $t-1$ can be calculated by:

$$N_{TD,r,t}^s = N_{DD,r,t}^s - N_{DD,r,t-1}^s \quad (17)$$

when $N_{TD,r,t}^s$ is larger than a certain threshold $N_{TD}^{threshold}$, the cycle slip occurs at time $t$. According to (16), the TD uses range distance to estimate the DD integer ambiguity, which relies on the high quality of the initial guess of the position. Thanks to LiDAR and IMU, the system can provide a high-quality initial guess, which even enables the detection of the small cycle slip.

*3) GNSS Doppler Factor*

Doppler measurements $d_{r,t}^s$ received by receiver $r$ from satellite $s$ at time epoch $t$ is denoted as:

$$\lambda d_{r,t}^s = \mathbf{e}_{r,t}^{s,LOS}(\mathbf{v}_{s,t}^{EC} - \mathbf{v}_{r,t}^{EC}) + c(\dot{\delta}_{r,t}^i - \dot{\delta}_{s,t}) + \varsigma_{r,t}^s \quad (18)$$

where $\varsigma_{r,t}^s$ represents the noisy term of the received Doppler measurement, c denotes the speed of light. $\lambda$ denotes the carrier wavelength of the respective satellite constellation system. The velocity of the receiver in the ECEF frame is transformed from the ENU frame through $\mathbf{v}_{r,t}^{EC} = \mathbf{R}_{EN}^{EC}\mathbf{v}_{r,t}^{EN}$. $\mathbf{e}_{r,t}^{s,LOS}$ is the LOS unit vector between the position of receiver $r$ and the satellite $s$ at time epoch $t$, which is calculated by:

$$\mathbf{e}_{r,t}^{s,LOS} = \left(\frac{\mathbf{p}_{s,t}^{EC} - \mathbf{p}_{r,t}^{EC}}{\|\mathbf{p}_{s,t}^{EC} - \mathbf{p}_{r,t}^{EC}\|}\right)^T \quad (19)$$

Given the Doppler measurement model above, the residual is derived as:

$$r_{\dot{d},r,t}^s = d_{r,t}^s - \frac{1}{\lambda_i}\left(\mathbf{e}_{r,t}^{s,LOS}(\mathbf{v}_{s,t}^{EC} - \mathbf{v}_{r,t}^{EC}) + c(\dot{\delta}_{r,t}^i - \dot{\delta}_{s,t})\right) \quad (20)$$

*4) Inertial Factor*

The IMU measurements contain linear acceleration and angular velocity with the effect of corresponding bias and additive noises. Knowing that the frequency of the inertial measurements is practically much higher than other sensors (LiDAR, GNSS), the pre-integration method [39] is further adopted in our optimization to integrate multiple raw inertial measurements into a single relative pose constraint between two consecutive keyframes $k$ and $k + 1$. We follow the work from [3] for the implementation. The readers can refer to [3, 39] for the detailed formulation of the inertial factors.

*5) LiDAR Landmark-Based Virtual Satellite Factor*

The satellite geometry is often poor in urban areas and will be further deteriorated by NLOS exclusion. As shown in Figure 5a, only the very limited LOS satellites (the blue circles) with high elevation angles remain after the GNSS NLOS exclusion.

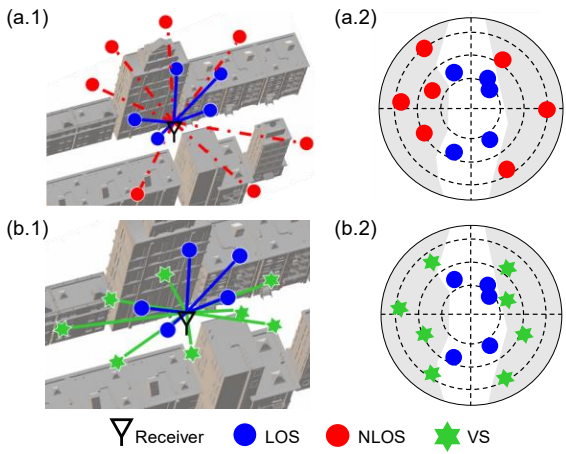

Fig. 5. The illustration of virtual satellite-aided GNSS-RTK positioning.

This would lead to significantly higher ADOP, limiting the performance of the integer ambiguity resolution. On the contrary, as shown in Figure 5b, the green stars denote the virtual satellites, which mainly arise from low-lying environmental structures and are highly complementary with high-elevation angle LOS satellites. Hence, this paper proposes to employ the virtual satellites to assist GNSS-RTK ambiguity resolution.

The employment of VS measurements and constraints is following a similar manner to the popular feature-based LiDAR-SLAM methods [10]. We have evaluated edge-based and plane-based LiDAR odometry in urban areas. The results showed that planar features obtained better accuracy and higher robustness. Therefore, planar constraints are applied in a scan-to-map scheme in this paper, where the map denotes a local 3D point cloud feature map accumulated by recent keyframes. For each keyframe, feature points of planes are extracted by evaluating the local distribution of the neighboring patch. Then the plane correspondences are found by nearest neighbor search between the keyframes and the local feature map and examined by eigenvalue analysis on the feature patches. Given the transformed planar point $\mathbf{p}_{p,k}^{EN}$ in frame $k$ and correspondent planar points $\mathbf{p}_{p,k,a}^{EN,M}$, $\mathbf{p}_{p,k,b}^{EN,M}$ and $\mathbf{p}_{p,k,c}^{EN,M}$ representing the planar patch in the local feature map $M$ in the ENU frame, the point-to-plane residual is calculated as [10]:

$$\mathbf{r}_{l,s,k} = \left\|\frac{\left(\mathbf{p}_{p,k}^{EN} - \mathbf{p}_{p,k,a}^{EN,M}\right)}{\left(\mathbf{p}_{p,k,a}^{EN,M} - \mathbf{p}_{p,k,b}^{EN,M}\right)\times\left(\mathbf{p}_{p,k,a}^{EN,M} - \mathbf{p}_{p,k,c}^{EN,M}\right)}\right\| \quad (21)$$

$$\mathbf{p}_{p,k}^{EN} = \mathbf{R}_{b,k}^{EN}(\mathbf{R}_l^b \mathbf{p}_{p,k}^l + \mathbf{p}_l^b) + \mathbf{p}_{b,k}^{EN} \quad (22)$$

where $\mathbf{p}_{p,k}^l$ represents the planar point in $k^{th}$ LiDAR keyframe, $\mathbf{T}_l^b = [\mathbf{R}_l^b \ \mathbf{p}_l^b]$ denotes the transformation matrix from the LiDAR frame to the body frame, $\mathbf{T}_{b,k}^{EN} = [\mathbf{R}_{b,k}^{EN} \ \mathbf{p}_{b,k}^{EN}]$ denotes the transformation matrix from $k^{th}$ local body frame to the ENU frame.

Moreover, to balance the impact of the virtual satellites and the original satellites in the optimization, only 200 virtual satellites are randomly selected for each keyframe. The lack of enough features reduces state observability, while the abundance of features results in excessive redundancy. The constraints constructed from virtual satellites are further dynamically weighted based on the quantity ratio between the numbers of virtual satellites and real satellites instantly, which is calculated by:

$$w_t^l = \frac{n_{virtual}^l}{n_{real}^l} \quad (23)$$

where $n_{virtual}^l$ represents the constraints' number of virtual satellites, $n_{real}^l$ represents the constraint number of real satellites. This enables real satellites to be effective even in small numbers, which is common in urban cases.

How is the satellite geometry improved by the VS factors? We can find support from the calculation of covariance. Given an optimization problem, the overall covariance matrix of the optimized system states $\chi^*$ are estimated as:

$$Cov_{\chi^*} = \left(\mathbf{J}_{\chi^*}^T \mathbf{W} \mathbf{J}_{\chi^*}\right)^{-1} \quad (24)$$

where $\mathbf{J}_{\chi^*}$ denotes the Jacobian matrix of the residuals in respect of the optimized system states. $\mathbf{W}$ represents the weighting matrix relative to different residuals. In this paper, by applying constraints from VS, IMU, and GNSS (pseudorange, carrier phase, and Doppler) the system observation function (discussed above) for the float solution $\chi_t^* = [\mathbf{p}_t^{EN}\ N_{DD,r,t}]$ is further summarized as:

$$\begin{bmatrix}\psi_{DD,t}\\ \rho_{DD,t}\\ d_t\\ L_t\\ B_t\end{bmatrix} = \begin{bmatrix}\lambda^{-1}\mathbf{G}_t^\psi & \mathbf{I}_{s\times s}\\ \mathbf{G}_t^\rho & \mathbf{0}\\ \mathbf{G}_t^d & \mathbf{0}\\ \mathbf{G}_t^L & \mathbf{0}\\ \mathbf{G}_t^B & \mathbf{0}\end{bmatrix}\begin{bmatrix}\mathbf{p}_t^{EN}\\ N_{DD,r,t}\end{bmatrix} + \begin{bmatrix}\varepsilon_{DD,r,t}^s\\ \epsilon_{DD,r,t}^s\end{bmatrix} \quad (25)$$

where $\psi_{DD,t}$, $\rho_{DD,t}$ and $d_t$ denote the measurements from double-differenced carrier phase, the double-differenced pseudorange, and the Doppler in time epoch $t$, respectively. $L_t$ and $B_t$ represent the measurements from LiDAR-based virtual satellites and pre-integrated IMU. $\mathbf{G}_t^{(\cdot)}$ represents the observation model of each measurement type to the states in the float solution. $\varepsilon_{DD,r,t}^s$ and $\epsilon_{DD,r,t}^s$ are the noise terms of position and integer ambiguities, respectively. Based on (24) and (25) the covariance matrix $\mathbf{Q}_{\chi_t^*}$ can be derived as:

$$\mathbf{Q}_{\chi_t^*} = \left(\begin{bmatrix}\lambda^{-1}\mathbf{G}_t^\psi & \mathbf{I}_{s\times s}\\ \mathbf{G}_t^\rho & \mathbf{0}\\ \mathbf{G}_t^d & \mathbf{0}\\ \mathbf{G}_t^L & \mathbf{0}\\ \mathbf{G}_t^B & \mathbf{0}\end{bmatrix}^T \mathbf{W}_t' \begin{bmatrix}\lambda^{-1}\mathbf{G}_t^\psi & \mathbf{I}_{s\times s}\\ \mathbf{G}_t^\rho & \mathbf{0}\\ \mathbf{G}_t^d & \mathbf{0}\\ \mathbf{G}_t^L & \mathbf{0}\\ \mathbf{G}_t^B & \mathbf{0}\end{bmatrix}\right)^{-1} \quad (26)$$

where the $\mathbf{W}_t'$ is the weighting matrix and consists of submatrices for the weights of the DD pseudorange factor $\mathbf{W}_t^\rho$, DD carrier phase factor $\mathbf{W}_t^\psi$, Doppler factor $\mathbf{W}_t^d$, VS factor $\mathbf{W}_t^L$ and inertial factor $\mathbf{W}_t^B$. It can be illustrated as:

$$\mathbf{W}_t' = \begin{bmatrix}\mathbf{W}_t^\psi & 0 & 0 & 0 & 0\\ 0 & \mathbf{W}_t^\rho & 0 & 0 & 0\\ 0 & 0 & \mathbf{W}_t^d & 0 & 0\\ 0 & 0 & 0 & \mathbf{W}_t^L & 0\\ 0 & 0 & 0 & 0 & \mathbf{W}_t^B\end{bmatrix} \quad (27)$$

The final result of $\mathbf{Q}_{\chi_t^*}$ can be derived as:

$$\mathbf{Q}_{\chi_t^*} = \begin{bmatrix}\mathbf{A} & \lambda^{-1}\mathbf{G}_t^{\psi T}\mathbf{W}_t^\psi\\ \lambda^{-1}\mathbf{G}_t^\psi\mathbf{W}_t^\psi & \mathbf{W}_t^\psi\end{bmatrix}^{-1} = \begin{bmatrix}\mathbf{Q}_{nn} & \mathbf{Q}_{np}\\ \mathbf{Q}_{pn} & \mathbf{Q}_{pp}\end{bmatrix} \quad (28)$$

With $\mathbf{A} = \lambda^{-2}\mathbf{G}_t^{\psi T}\mathbf{W}_t^\psi\mathbf{G}_t^\psi + \mathbf{G}_t^{\rho T}\mathbf{W}_t^\rho\mathbf{G}_t^\rho + \mathbf{G}_t^{d T}\mathbf{W}_t^d\mathbf{G}_t^d + \mathbf{G}_t^{L T}\mathbf{W}_t^L\mathbf{G}_t^L + \mathbf{G}_t^{B T}\mathbf{W}_t^B\mathbf{G}_t^B$

where $\mathbf{Q}_{nn}$ and $\mathbf{Q}_{pp}$ denotes the variance matrices of the float integer ambiguities and position results, respectively. $\mathbf{Q}_{np} = \mathbf{Q}_{pn}^T$ denotes the covariance matrices of the float integer ambiguities and position results. Compared with the conventional GNSS-only and GNSS/INS methods, the $\mathbf{Q}_{np}$, $\mathbf{Q}_{pn}$ and $\mathbf{Q}_{pp}$ remain the same. However, the $\mathbf{Q}_{nn}$ depends not only on the measurements from surviving limited satellites after NLOS exclusion but also on the substantial impact of VS constraints. To intuitively demonstrate effectiveness, the analysis of ADOP is adopted to evaluate the success rate of AR. The ADOP calculation is in the following form:

$$ADOP = \sqrt{det(\mathbf{Q}_{nn})^{\frac{1}{m}}} \quad (29)$$

where $det(\cdot)$ represents the determinant calculation. It is known that the higher success rate of AR is with a lower value of ADOP. In the following section, the effect of the virtual satellite measurements on the covariance matrix to the integer ambiguity states and thus on the AR success rate will be experimentally demonstrated and evaluated.

Given the factors above, the optimized states and their respective covariance matrix are obtained after the tightly-coupled optimization. Knowing that the double-differenced integer ambiguities should indeed be integer values, the estimated float values of integer ambiguities can be resolved as integer values and the position results can be corrected with higher accuracy, whereas a fixed solution is reached. To solve the integer ambiguity resolution problem, the LAMBDA algorithm [17] is adopted by searching for the integer solution based on the improved float solution in the searching space defined by the correspondent covariance matrix.

*6) Marginalization Factor*

To release the computational load and meanwhile maintain the impacts of the constraints from the previous information, marginalization is adopted in the sliding window optimization. We gradually marginalize the constraints from the older keyframes sliding out the window through the Schur complement [43]. The corresponding new prior factor is further added in the updated window.

## IV. EXPERIMENTAL RESULTS

The proposed system is implemented using C++ on Robot Operation System (ROS) [44]. We use Ceres Solver [45] and GTSAM [46] for the nonlinear optimization as well as the pose graph optimization. The experimental evaluation is conducted on two challenging sequences in an open-source dataset, *UrbanNav* [47], which contains various levels of urbanized scenarios. Both the two sequences are collected in typical urban canyons, where densely distributed static buildings, tall trees, and dynamic objects (wagons, double-decker buses) bring potential GNSS NLOS receptions. In the first sequence (denoted as urban canyon 1 in Figure 6), the height and the density of the buildings are lower than the buildings in the second sequence (denoted as urban canyon 2 in Figure 13).

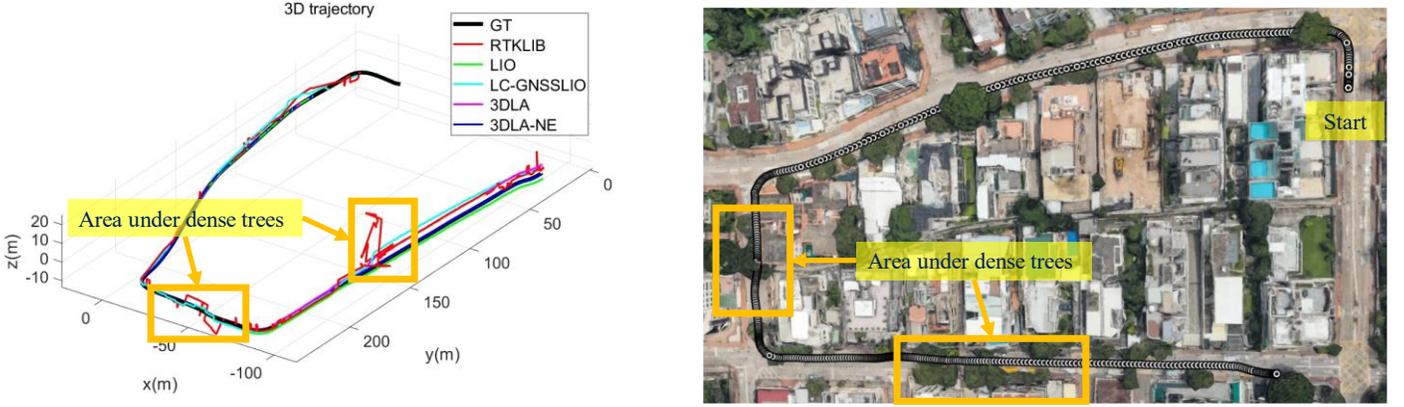

Fig. 6. The trajectory in Urban Canyon 1. Left side shows the 3D trajectories of different methods. The red, green, cyan, magenta, and blue curves denote the RTK, LIO, LC-GNSSLIO, 3DLA, and 3DLA-NE, respectively. The black curve denotes the ground truth trajectory. Right side shows the projected trajectories in Google Earth. Orange box denotes the area with challenging signal blockage by trees.

## A. Experiment Platform

The data collection is based on the experimental platform proposed in *UrbanNav*. A low-cost GNSS receiver, the u-blox M8T, is employed to collect raw single-frequency GPS/BeiDou signals at 10 Hz. The Xsens Ti-10 IMU is adopted to collect inertial measurements at the frequency of 100 Hz. The HDL-32E collects 3D measurements at a frequency of 10 Hz. Moreover, the NovAtel SPAN-CPT, an integration system from multiple frequencies and constellation GNSS-RTK and IMU with fiber-optic gyroscopes (FOG, 1 degree per hour for gyro bias, 0.067 degrees per hour as random walk), is employed to provide ground truth of positioning. Noted that the baseline between the receiver and the GNSS base station is less than 7km and with the help of the post-processing software from NovAtel, the absolute accuracy of the ground truth could be guaranteed. In the implementation, data from different sources are synchronized via Paulse Per Second (PPS) hardware in ROS. The extrinsic parameters between different sensors are carefully calibrated before the experiments. The initial transformation from local coordinate to global coordinate is provided in advance by aligning the first position to ground truth.

## B. Evaluation Comparison

To evaluate the effectiveness of the proposed method, the following methods are evaluated and compared qualitatively and quantitatively in multiple aspects. First, mean error, maximum error, and standard deviation in both 2D and 3D cases will demonstrate the positioning accuracy of different methods. Second, the fix rate can indicate the effectiveness of geometry improvement by the proposed method. Moreover, the availability is also evaluated to compare the positioning ability. The evaluated methods are listed as follows:

(a) **RTK**: RTKLIB [42] is adopted to represent the performance of the conventional GNSS-RTK. Forward filtering is adopted under fix-and-hold conditions.
(b) **LIO**: LiDAR/Inertial integration method in [11] is evaluated to demonstrate the performance.
(c) **LC-GNSSLIO**: Loosely-coupled (LC) integration between GNSS-RTK and LIO system. The method in [48] is performed to show the improvement of positioning by loosely integrating the GNSS-RTK with the LIO system.
(d) **3DLA**: The proposed tightly-coupled VS-aided GNSS-RTK/IMU integrated system. This is to show the effectiveness of geometry improvement by VS.
(e) **3DLA-NE**: The proposed tightly-coupled VS-aided GNSS-RTK/IMU integrated system with drift-free NLOS exclusion. This is to demonstrate the final performance of the proposed method.

*1) Evaluation of Urban Canyon 1*

*a) Evaluation of the positioning performance*

Table. 1 shows the evaluation results of each method. The trajectory and the 3D positioning error are illustrated in Figure 6 and Figure 7. The GNSS-RTK has shown the result with 1.55 meters as the 2D mean error and 3.54 meters as the 3D mean error. Its maximum error reaches 15.88 meters in 2D and 29.23 meters in 3D. By analyzing the position where the large error occurs (the orange box in Figure 6 corresponds to time intervals around the 70s and 120s in Figure 7), it is found that these spots all suffer from dense buildings and trees, which caused severe NLOS receptions with greatly deteriorated accuracy. Moreover, the conventional GNSS-RTK method finally achieves a fix rate of 14.01%.

Table. 1. Positioning performance of the evaluated five methods in urban canyon 1. 2D MEAN/3D MEAN represents horizontal and 3D positioning errors in meters. The improvement (Impr.) is calculated concerning the RTK method. STD denotes the standard deviation. "Avail." denotes the availability.

| ALL DATA | GNSS-RTK | LIO | LC-GNSSLIO | 3DLA | 3DLA-NE |
|---|---|---|---|---|---|
| 2D MEAN | 1.55 | **0.32** | 0.83 | 0.39 | 0.36 |
| 2D MAX | 15.88 | 0.97 | 2.67 | 0.76 | **0.83** |
| 2D STD | 1.32 | 0.23 | 0.48 | 0.16 | **0.16** |
| 2D IMPR. | | **79.35%** | 46.45% | 74.84% | 76.77% |
| 3D MEAN | 3.54 | 1.30 | 2.90 | 1.53 | **0.44** |
| 3D MAX | 29.23 | 2.79 | 6.34 | 5.29 | **0.87** |
| 3D STD | 3.38 | 0.81 | 1.36 | 1.78 | **0.15** |
| 3D IMPR. | | 63.28% | 18.08% | 56.78% | **87.57%** |
| FIXED RATE | 14.01% | | 14.01% | 20.91% | **31.37%** |
| AVAIL. | 76.6% | 100% | 100% | 100% | **100%** |

On the other hand, the LIO achieved a mean error of 0.32 meters in 2D and 1.30 meters in 3D. However, it suffers from drift with increasing driving distance and ends with a maximum 3D error of 2.79 meters. We further refer to the result of the loosely-coupled method and notice the improvement in the positioning accuracy, as LIO can help GNSS-RTK with accurate relative pose constraints to provide a smoother, globally accurate positioning result. The accuracy improvement can be observed by decreasing 2D and 3D errors by 0.83 meters and 2.90 meters, better smoothness is observed with lower maximum error and standard deviation. Nevertheless, the positioning error is still quite large where severe signal blockages occur. In other words, the loosely-coupled method can cope with the drift problem in long-term operation but the local positioning accuracy is greatly affected by the GNSS-RTK result. Therefore, the improvement is still limited by the unhealthy GNSS measurements and underutilized complementary characteristics of GNSS and LiDAR/IMU.

3DLA, as a tightly-coupled system integrating the raw measurements from GNSS DD pseudorange, DD carrier phase, and Doppler with virtual satellite and IMU measurements, shows significant improvement compared with conventional GNSS-RTK method and its loosely-coupled integration method with LIO: The 2D error and 3D error decrease to 0.39 and 1.53 meters, whereas the standard deviation and maximum error are 1.78 and 5.29 meters in 3D case. Firstly, compared with loosely-coupled integration through relative pose, the tightly-coupled integration fuses all the raw measurements that can precisely model the optimization problem and mitigate the impact of outliers. Secondly but more importantly, the virtual satellite measurements make a great contribution to improving the original satellite geometry. The fix rate of 3DLA reaches 20.91%, yet the potential NLOS receptions are still not excluded. Therefore, 3DLA-NE is performed to further demonstrate the effectiveness of the proposed NLOS exclusion method. 3DLA-NE eventually shows the best performance with 0.36 meters and 0.44 meters as 2D and 3D error, 0.15 meters and 0.87 meters as 3D standard deviation and 3D maximum error. More importantly, the fix rate of the whole trajectory by 3DLA-NE reaches 31.37%. These two observations demonstrate that the NLOS exclusion and the geometry improvement through the virtual satellites can make significant contributions to the final positioning result.

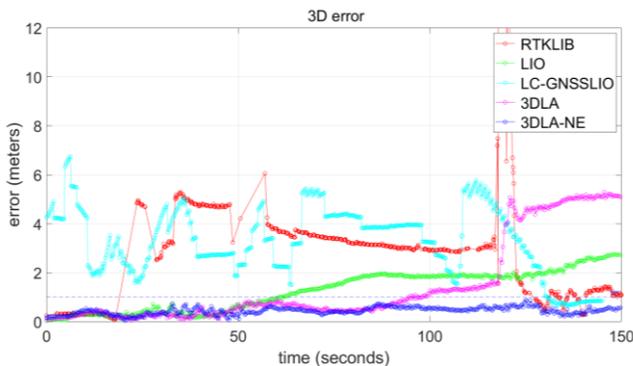

Fig. 7. 3D Positioning error in the urban canyon 1. The red, green, cyan, magenta, and blue curves denote the RTK, LIO, LC-GNSSLIO, 3DLA, and 3DLA-NE, respectively

*b) Analysis of NLOS detection*

Figure 8 illustrates how the drift error of attitude estimation will mislead the NLOS classification. The left figure shows the result of NLOS detection based on the global optimized drift-free PCM while the right figure represents the results on drifted PCM. The accumulated attitude drift significantly alters actual satellite occlusion, preventing correct detection of NLOS receptions. Furthermore, the positioning error of the above two cases is compared in Table 2, the method with accumulated drift shows higher error, which is because some NLOS receptions are not detected, and healthy measurements are mistakenly classified as NLOS and excluded.

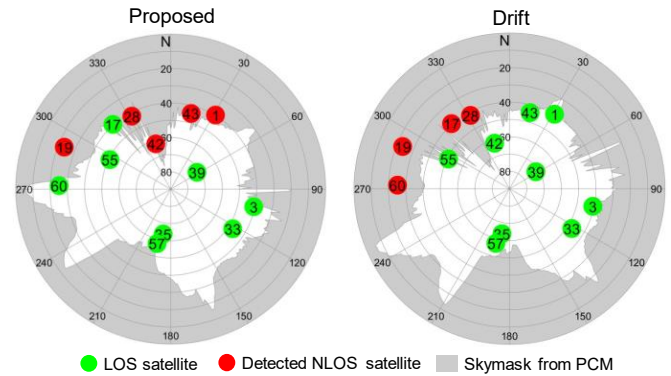

Fig. 8. Comparison between the NLOS detection methods based on the globally optimized drift-free PCM (left) and the drifted PCM (right). The skymasks generated from different PCM are shown in gray.

Moreover, we have also evaluated the proposed method with different sliding window lengths to demonstrate the ability of the PCM to reconstruct the environment and thus aid NLOS detection. Figure 9 shows the comparison of NLOS detection through the PCM generated by the different sliding window sizes. As the window size in the left figure is 60 keyframes as default and in the right figure is 20 keyframes, it is apparent that the upper part of the building has been more thoroughly rebuilt in the left figure than in the right one. In other words, the larger sliding window size represents a more complete reconstruction of the environment, thus allowing a more accurate determination of the satellite occlusion. Additionally, the evaluation results are shown in Table 2.

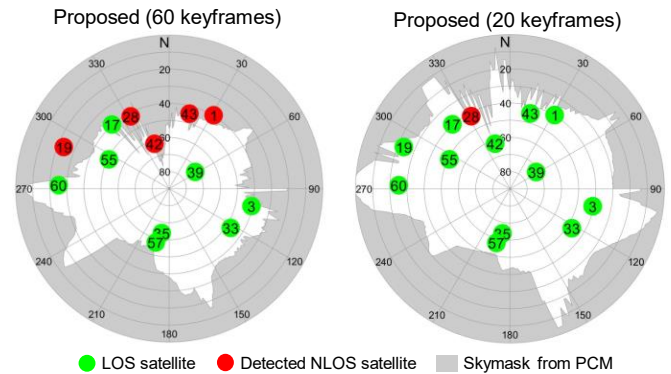

Fig. 9. Comparison between the results of NLOS detection methods based on different sliding window sizes.

Table. 2. Positioning performance (meters) of the evaluated three cases for the selected epoch in urban canyon 1.

| All data | Drift PCM | Proposed (20 keyframes) | Proposed (60 keyframes) |
|---|---|---|---|
| **3D Error (m)** | 1.96 | 1.17 | **0.55** |

*c) Analysis of Cycle Slip Detection*

To guarantee the constraints of the carrier phase measurements are constructed correctly, we apply LiDAR-aided cycle slip detection following the method in [40] and assess its effectiveness. Figure 10 illustrates the result of cycle slip detection by the LiDAR-aided method and by receiver indicator. At most of the epochs, the proposed method detects more cycle slip than the original receiver indicator, the sum of the cycle additionally detected by the LiDAR-aided method reaches more than 10 cycles. Moreover, the positioning error of the proposed 3DLA GNSS-RTK method (denoted as 3DLA-NE) and the proposed method without LiDAR-aided cycle slip detection (denoted as 3DLA-CL) are evaluated to verify the validity of the detected cycle slip. It is observed that without LiDAR-aided cycle slip detection, the deteriorated positioning accuracy denotes that the cycle slip is not sufficiently detected.

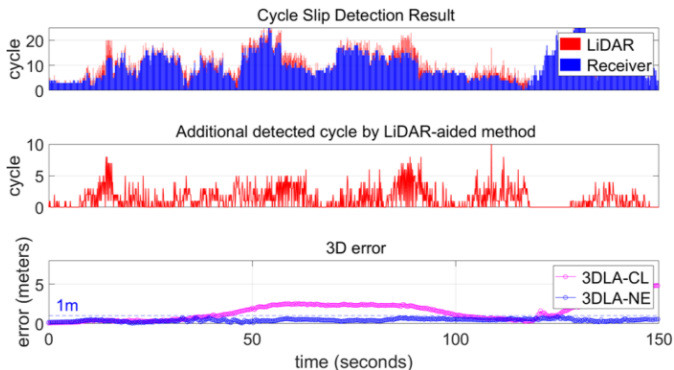

Fig. 10. The figure at the top shows the result of cycle slip detection by different methods. The blue bar denotes the cycle slip number detected by the receiver indicator. The red bar denotes the number of detected cycle slips by the LiDAR-aided method. The figure in the middle indicates the sum of the additionally detected cycle slips by the LiDAR-aided method. The figure at the bottom shows the positioning accuracy of different methods.

*d) Analysis of Geometry Improvement*

As discussed in Section III, the effectiveness of the proposed virtual satellite-aided integer ambiguity resolution is indicated by the ADOP value. To better demonstrate the contribution of the virtual satellites, the ADOP value from methods with different weighting on virtual satellites is presented in Figure 11. It can be observed that the 3DLA GNSS-RTK shows a significantly decreased ADOP value compared with the conventional GNSS-RTK method with the help of virtual satellites. Moreover, there is a notable downward trend in the ADOP value as the weight of virtual satellites increases. The observation proves that the virtual satellites effectively improve the original geometry, yielding more accurate positioning results and a higher fix rate. As shown in Figure 12, the proposed 3DLA GNSS-RTK outperforms the conventional GNSS-RTK method by achieving more fix solutions where the conventional GNSS-RTK fails.

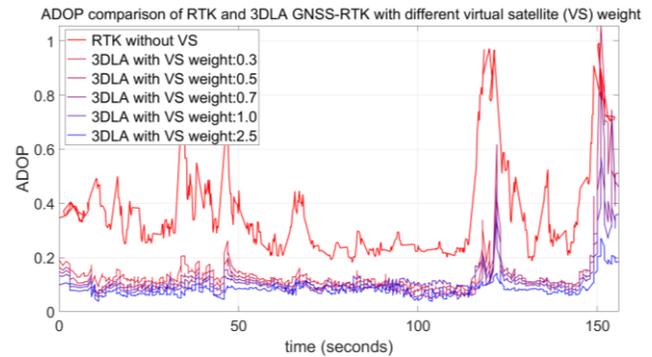

Fig. 11. ADOP value by different weighting of virtual satellite (denoted as VS). From red to blue, the color denotes the weighting from 0.0 to 2.5, respectively.

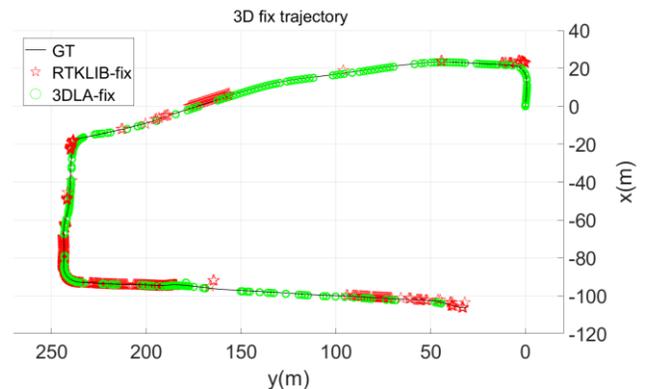

Fig. 12. Results of the fixed solutions on the trajectory for Urban Canyon 1. The red stars represent the fixed solutions obtained by the conventional GNSS-RTK method. The green dots represent the fixed solutions obtained by the proposed 3DLA GNSS-RTK.

In conclusion, the proposed 3D LiDAR-aided GNSS-RTK/IMU integrated positioning method firstly filters out unhealthy GNSS measurements from potential NLOS receptions based on precisely modeled PCM, and secondly improves the satellite geometry by virtual satellites from LiDAR measurements, which leads to better positioning accuracy, higher fix rate, and higher robustness. The effectiveness of two key contributions can be observed through the gradually changed error curves in Figure 7. One should be noted that the maximum remaining error still reaches 0.87 meters, which can be inferred as the result of multipath receptions, which cannot be detected directly through the PCM. How to infer the uncertainty of the multipath effects from the 3D PCM is also an interesting topic to be investigated which is one of our future directions.

*2) Evaluation of Urban Canyon 2*

The experimental evaluation is also conducted on Urban Canyon 2 to show the effectiveness of the proposed method applied in a more urbanized area. Table 3 demonstrates the results of the compared methods, Figure 13 and Figure 14 illustrate the 3D trajectory and 3D positioning error. Urban Canyon 2 is more challenging for positioning compared with Urban Canyon 1 due to more buildings and trees. The GNSS-RTK achieves 1.81 meters of 2D mean error and 3.65 meters of 3D mean error. The maximum 3D error reaches 55.59 meters with 5.27 meters as the standard deviation. The overall fix rate is 3.45%. Moreover, the LIO shows a better performance with 1.76 meters of 2D error and 1.97 meters of 3D error. It is observed in Figure 14 that the increasing positioning error represents the accumulated drift. Furthermore, the LC-GNSSLIO achieves 1.38 meters mean error in the 2D case and 2.77 meters in the 3D case. Although the loose integration between LIO and GNSS-RTK helps to obtain a better positioning result, the impact of the NLOS receptions and non-ideal geometry distribution are still not compensated.

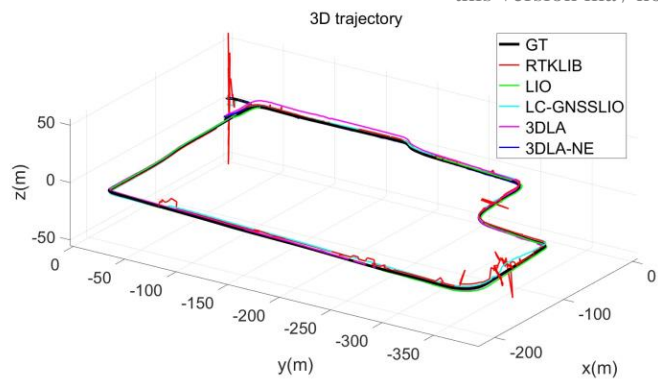
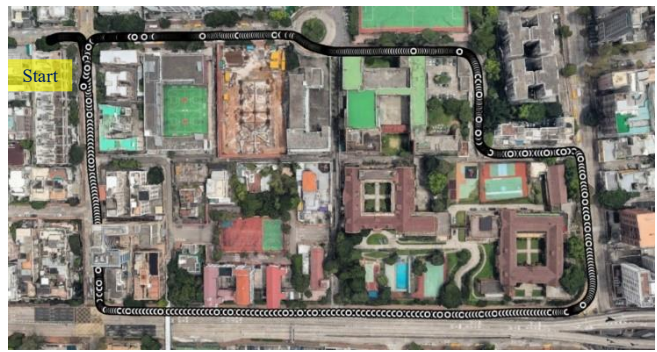

Fig.13. The trajectory in Urban Canyon 2. Left side shows the 3D trajectories of different methods. The red, green, cyan, magenta, and blue curves denote the RTK, LIO, LC-GNSSLIO, 3DLA, and 3DLA-NE, respectively. The black curve denotes the ground truth trajectory. Right side shows the projected trajectories in Google Earth.

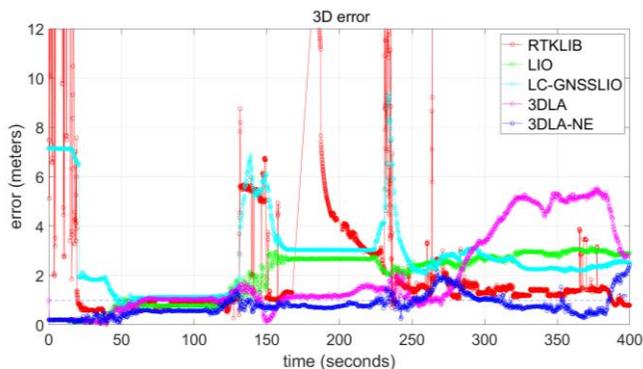

Fig. 14. 3D Positioning error in Urban Canyon 2.

Fortunately, the geometry problem can be properly tackled by the proposed tightly-coupled virtual satellite-aided GNSS-RTK manner. 3DLA shows an increase of 2D mean error as 0.61 meters and a fix rate of 12.93%. However, the 3D mean error reaches 2.02 meters as the remaining NLOS receptions are not excluded. Therefore, 3DLA-NE with accurate NLOS exclusion is finally performed and achieves the best accuracy with 2D mean error and 3D mean error decreasing to 0.49 meters and 0.79 meters respectively. Meanwhile, the maximum error and the standard deviation in 2D and 3D cases are 1.44/0.22 meters and 2.16/0.42 meters. More importantly, the fix rate of the 3DLA-NE reaches the highest level at 19.19%. Figure 15 shows that more fix solutions are obtained by the proposed approach compared with the conventional GNSS-RTK method. The evaluation of urban canyon 2 further proves the effectiveness of the proposed method of two key contributions as LiDAR-aided NLOS exclusion and geometry improvement.

Table. 3. Positioning performance of the evaluated five methods in urban canyon 2. 2D MEAN/3D MEAN represents horizontal and 3D positioning errors in meters. The improvement (Impr.) is calculated concerning the RTK method. STD denotes the standard deviation. "Avail." denotes the availability.

| ALL DATA | GNSS-RTK | LIO | LC-GNSSLIO | 3DLA | 3DLA-NE |
|---|---|---|---|---|---|
| 2D MEAN | 1.81 | 1.76 | 1.38 | 0.61 | **0.49** |
| 2D MAX | 47.28 | 3.15 | 6.74 | 1.54 | **1.44** |
| 2D STD | 2.20 | 0.95 | 1.21 | 0.36 | **0.22** |
| 2D IMPR. | | 2.76% | 23.75% | 68.72% | **74.87%** |
| 3D MEAN | 3.65 | 1.97 | 2.77 | 2.02 | **0.79** |
| 3D MAX | 55.59 | 3.15 | 9.39 | 5.55 | **2.16** |
| 3D STD | 5.27 | 0.97 | 1.58 | 1.70 | **0.42** |
| 3D IMPR. | | 46.02% | 24.11% | 44.65% | **78.36%** |
| FIXED RATE | 3.45% | | 3.45% | 12.93% | **19.19%** |
| AVAIL. | 94.7% | 100% | 100% | 100% | **100%** |

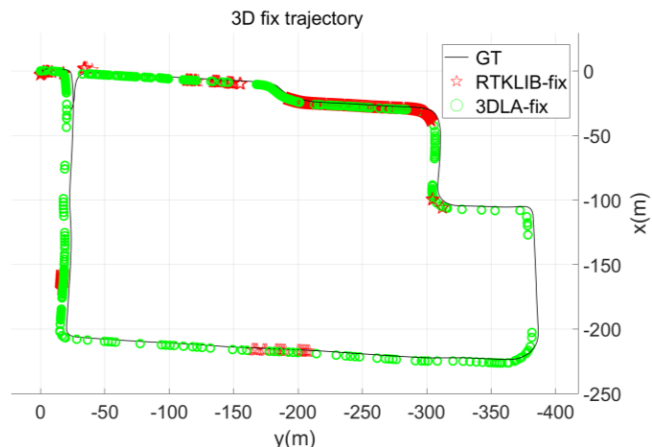

Fig. 15. Results of the fixed solutions on the trajectory for Urban Canyon 2.

## V. CONCLUSION

GNSS-RTK can provide reliable accurate positioning results in the opening area yet will suffer from NLOS receptions and poor satellite distribution in urban canyons, this paper presents 3DLA as 3D LiDAR-aided GNSS-RTK positioning that: (1) performs NLOS detection and exclusion based on the drift-free 3D PCM to eliminate the impact of unhealthy GNSS measurements, and (2) improves the satellite geometry distribution through the low-lying virtual satellites provided by LiDAR landmarks. The experiment results evaluated in two challenging sequences in Hong Kong shows that the proposed system can achieve higher accuracy and robustness in a highly urbanized area using commercial-level GNSS receivers and LiDAR/IMU sensor kit.

The main part of the remaining positioning errors can be attributed to multipath receptions. Different from NLOS receptions, multipath receptions are more difficult to identify through LiDAR and low-cost GNSS receivers. Therefore, one possible aspect of future work can focus on the detection and correction of multipath receptions. By tackling the potential NLOS receptions first, we can refer to the residuals during the optimization to detect the multipath receptions. On the other hand, the current 3DLA GNSS-RTK system is built in a sliding window optimization manner, which limits the potential of exploring the joint positioning ability of the GNSS measurements from more epochs. Given the high accuracy relative positioning ability of the LiDAR/IMU system, a method that guarantees the global consistency of both GNSS and LiDAR constraints based on a much larger sliding window would be a promising way to further enhance the positioning accuracy in a harsh urban canyon.

## APPENDIX

### A. Frame Definitions

Considering that multiple sensors fusion involves different spatial frames, the associated frames are defined as follows:

1) Earth-centered, earth-fixed (ECEF) frame [16]: The ECEF frame $(\cdot)^{EC}$ is a Cartesian coordinate system which is aligned with the center of the earth. It is typically used to express a satellite's position and related measurements.

2) East, north, and up (ENU) frame [16]: The ENU frame $(\cdot)^{EN}$ is a coordinate whose x, y, and z-axis align with east, north, and up directions. It is determined by a tangent plane to the surface of the earth and used to connect the local world frame with the ECEF frame. It should be noted that in this paper all the related states are transformed in the ENU frame by default.

3) Sensor frame: The sensor frame is a coordinate attached to local sensors. Sensor frames involved in this paper are denoted with $(\cdot)^b$, $(\cdot)^l$, $(\cdot)^r$ respect to IMU, LiDAR, and GNSS receiver frames.

### B. Notations

In this paper, uppercase bold letters are employed to denote matrices, and lowercase bold letters are used for vectors. Variables and frame coordinates are denoted as italic letters and constant scalars are denoted as lowercase letters. For the rest of the paper, major notations are defined below:

1) The satellite pseudorange measurement is denoted as $\rho_{r,t}^s$, where $r$ represents the GNSS receiver, $t$ represents the time index and $s$ denotes the satellite's index;

2) The carrier phase measurement received by receiver $r$ from satellite $s$ at time epoch $t$ is represented as $\psi_{r,t}^s$;

3) The Doppler measurement received by receiver $r$ from satellite $s$ at time epoch $t$ is represented as $d_{r,t}^s$;

4) The position and velocity of the satellite $s$ at time epoch $t$ are expressed by $\mathbf{p}_{s,t}^{EC}$ and $\mathbf{v}_{s,t}^{EC}$, respectively;

5) The position and velocity of the receiver $r$ at time epoch $t$ are expressed by $\mathbf{p}_{r,t}^{EN}$ and $\mathbf{v}_{s,t}^{EN}$, respectively;

6) The clock bias of the receiver $r$ on satellite constellation system $i$ at time epoch $t$ is expressed by $\delta_{r,t}^i$, the receiver clock bias drift is $\dot{\delta}_{r,t}^i$. The clock bias of the satellite $s$ at time epoch t is expressed by $\delta_{s,t}$, the satellite clock bias drift is $\dot{\delta}_{s,t}$. Note that the multiple satellite systems share the same receiver clock bias drift;

7) The position of the base (reference) station is expressed by $\mathbf{p}_e^{EC}$. The pseudorange and carrier phase measurements received by base station $e$ from satellite $s$ at time epoch $t$ are expressed by $\rho_{e,t}^s$ and $\psi_{e,t}^s$;

8) The origin point which connects the ECEF frame and ENU frame is denoted by $\mathbf{p}_o^{EC}$;

9) The transformation from frame $A$ to frame $B$ is expressed by $\mathbf{T}_A^B = [\mathbf{R}_A^B \ \mathbf{p}_A^B]$, where $\mathbf{R}_A^B$ denotes the rotation matrix with $\mathbf{q}_A^B$ as its quaternion form and $\mathbf{p}_A^B$ denotes the translation vector.

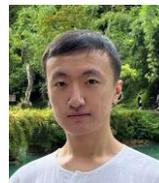

**Xikun Liu** received his bachelor's degree in Mechanical Design, Manufacturing, and Automation from Huazhong University of Science and Technology, China in 2017, and master's degree in Mechatronics and Information Technology from Karlsruhe Institute of Technology, Germany in 2021. He is currently pursuing a Ph.D. in the Department of Aeronautical and Aviation Engineering, the Hong Kong Polytechnic University. His research interests include GNSS and sensor-aided GNSS positioning, SLAM, and multiple sensor fusion in autonomous driving.

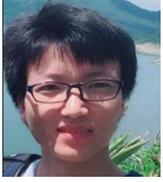
**Weisong Wen** was born in Ganzhou, Jiangxi, China. He received a Ph.D. degree in mechanical engineering, the Hong Kong Polytechnic University. He was a visiting student researcher at the University of California, Berkeley (UCB) in 2018. He is currently a research assistant professor in the Department of Aeronautical and Aviation Engineering. His research interests include multi-sensor integrated localization for autonomous vehicles, SLAM, and GNSS positioning in urban canyons.

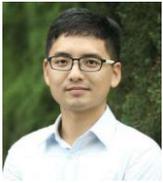
**Feng Huang** received his bachelor's degree from Shenzhen University in Automation in 2014 and MSc in Electronic Engineering at Hong Kong University of Science and Technology in 2016. He is a Ph.D. student in the Department of Aeronautical and Aviation Engineering, Hong Kong Polytechnic University. His research interests including localization and sensor fusion for autonomous driving.

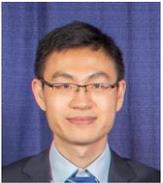
**Han Gao** received his Ph.D. degree from University College London (UCL) in 2019. He received a BSc degree in Aerospace Engineering from Shanghai Jiao Tong University (SJTU) in 2014. He is currently a senior software engineer at ZEEKR Technologies in China. He is interested in machine learning, localization, and mapping techniques for the autonomous system.

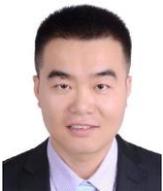
**Yongliang Wang** received the M.E. degree in the School of Electronic and Information Engineering from Xi'an Jiaotong University, China, in 2010. He is expertised in the field of wireless signal and image processing. He is currently the chief positioning technology engineer in Riemann Lab, 2012 Laboratories, Huawei Technologies Co., Ltd.

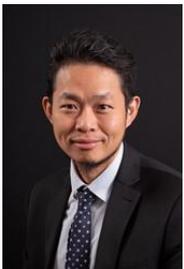
**Li-Ta Hsu** received B.S. and Ph.D. degrees in aeronautics and astronautics from National Cheng Kung University, Taiwan, in 2007 and 2013, respectively. He is currently an associate professor with the Department of Aeronautical and Aviation Engineering. The Hong Kong Polytechnic University, before he served as a post-doctoral researcher at the Institute of Industrial Science at the University of Tokyo, Japan. In 2012, he was a visiting scholar at University College London, the U.K. His research interests include GNSS positioning in challenging environments and localization for pedestrians, autonomous driving vehicle, and unmanned aerial vehicle.